# Challenges and Opportunities in Exoskeleton-based Rehabilitation

Rana Soltani-Zarrin, Amin Zeiaee, Reza Langari, Reza Tafreshi

*Abstract*— Robotic systems are increasingly used in rehabilitation to provide high intensity training for patients with motor impairment. The results of controlled trials involving human subjects confirm the effectiveness of robot-enhanced methods and prove them to be marginally superior over standard manual therapy in some cases. Although very promising, this line of research is still in its infancy and further studies are required to fully understand the potential benefits of using robotic devices such as exoskeletons. Exoskeletons have been widely studied due to their capability in providing more control over paretic limb as well as the complexities involved in their design and control. This paper briefly discusses the main challenges in development of rehabilitation exoskeletons and elaborates more on how some of these issues are addressed in the design of CLEVERarm, a recently developed upper limb rehabilitation exoskeleton. The paper is concluded with several remarks on the current challenges in wide-spread use of exoskeletons in medical facilities, and a vision for the future of these technologies in rehabilitation medicine.

## I. INTRODUCTION

Stroke is the leading cause of motor disabilities around the globe. Every year, 15 million people worldwide suffer a stroke. More than 85 percent of them survive, but only 10 percent recover completely [1]. The rest must deal with mobility impairments in upper and/or lower limb, cognitive disabilities or many other types of post stroke conditions. Stroke victims can get help relearning skills they have lost or learn new ways of performing tasks to compensate for lost abilities through Occupational Therapy (OT). Similarly, Physical Therapy (PT) can help stroke victims by reducing the muscle spasticity and pain, and improving their range of motion in the impaired joints.

The most effective rehabilitation is specific to the skills the patient needs, and of sufficiently high intensity and duration to truly retrain the nerves and muscles involved [2, 3]. However, there are limitations on the available resources such as the number of trained human therapists, while the demand is growing, particularly as population age. The U.S. Census Bureau estimates that the number of Americans age 65 or over, whom according to stroke research studies are at greater risk of suffering a stroke, will double by 2050 [4]. Resultantly, the number of occupational therapy and physical therapy jobs is expected to increase 27 percent and 34 percent, respectively, by 2020 according to the U.S. Bureau of Labor Statistics [5]. Though interest in the field is growing, the American Academy of Physical Medicine and Rehabilitation projects the current physical therapist shortage will increase significantly in the upcoming decades.

The inherent capabilities of robotics systems in producing highly repetitive, and precisely controllable motions make them desirable for rehabilitation purposes [6]. End effector based systems [7, 8] and exoskeletons [9-11] are the two main categories of robotic systems designed to provide automated therapy to stroke patients. While end effector based systems precede exoskeletons, the results of studies have proven that the latter category outperforms the former by providing more control over the motion of paretic limb [12]. Despite the advantages, there are major issues associated with the kinematic compatibility of exoskeletons with human arm, which makes design of prosthetic devices challenging. In addition to design issues, development of effective control algorithms appropriate for rehabilitation goals has been a major challenge [13]. Partially due to the aforementioned reasons, many exoskeletal systems are bulky, very complex to operate, costly and heavy which limit wide-spread use of them.

CLEVERarm, is a recently developed upper limb rehabilitation exoskeleton with 8 degrees of freedom supporting the motion of shoulder girdle, Glenohumeral joint, elbow and wrist [14]. The mechanical design of the exoskeleton is centered on reducing the weight and bulkiness of the whole structure. This paper gives a brief overview of the key features and functionalities of CLEVERarm, followed by a thorough discussion about the opportunities and challenges in use of exoskeletons in medical facilities. The paper is concluded with author's vision for the future of these technologies in rehabilitation medicine.

## II. CLEVERARM

CLEVERarm has six active, and two passive degrees of freedom, allowing the motion of shoulder girdle, glenohumeral (GH) joint, elbow, and wrist. An active degree of freedom (DOF) is used for assisting Flexion/Extension of the elbow, while the remaining five active degrees of freedom are used in the design of the device shoulder to improve the ergonomics of the device. Having five degrees

This research was supported by NSF grant 1644743 and QNRF, NPRP 7-1685-2-626.

R. Soltani Zarrin is with Texas A&M University, College Station, TX, USA (e-mail: rana.soltani@ tamu.edu).
A. Zeiaee and R. Langari are with Texas A&M University, College Station, TX, USA (email:amin.zeiaee@tamu.edu, rlangari@tamu.edu).
R. Tafreshi is with Texas A&M University at Qatar, Doha, Qatar (reza.tafreshi@qatar.tamu.edu)

of freedom, the proposed shoulder design supports the 2D motion of GH joint center on the body frontal plane. The two passive DOF of CLEVERarm allow the pronation/supination, and flexion/extension motions of the wrist. Figure 1 shows the CAD model of the CLEVERarm:

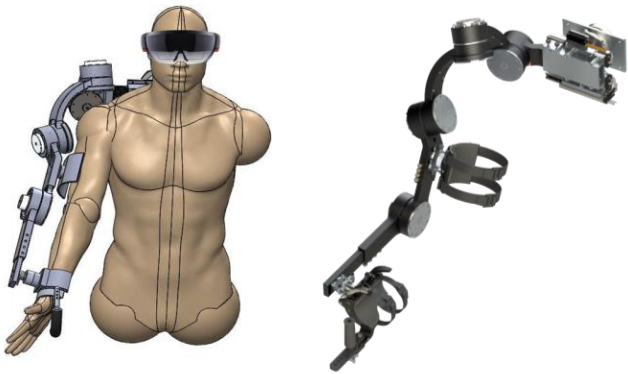

Fig 1. CLEVERarm CAD Model

Design of CLEVERarm is focused on addressing two important shortcomings common to many exoskeletons: weight and bulkiness. The number of degrees of freedom used to model biologic joints is an important factor affecting the complexity of the structure. The motion of biologic joints is in general very hard to replicate with one DOF hinged joint. However, using higher number of degrees of freedom to model these motions results in bulky and complex systems. For example, NEUROExos elbow exoskeleton uses 4 passive DOF in a complex passive mechanism to support the full biologic motion of the elbow joint [15]. For joints supporting more than one dominant rotation (e.g. shoulder and wrist), it is also challenging to design links connecting the consecutive degrees of freedom while preserving the natural range of motion of the joint by avoiding possible collision between the device and body. The choice of the consecutive axes of rotations in the shoulder mechanism of CLEVERarm is to minimize the volume while achieving the three degrees of freedom of Glenohumeral joint.

Actuation units, gearing system and the metal body of the robotic devices are in general not lightweight. Therefore, exoskeletons are usually heavy mechanisms. To minimize the weight of the device body, various choices of material and manufacturing techniques were studied while taking into consideration the cost and feasibility of the mass production. A combination of Aluminum and 3D printed Carbon Fiber (CF) reinforced plastic were chosen as the body material for the exoskeleton. While the use of CF for rehabilitation robotic devices has been suggested in the past [10, 16], to the best of our knowledge no actual realization of such an exoskeleton have been reported. This might be partly due to the non-isotropic properties of composite materials which make the design of a CF robotic device that undergoes various loading scenarios challenging. On the other hand, certain components of the system such as connectors for the motor shafts and fasteners cannot be made from reinforced plastic and the design of metal/plastic interfaces proves to be a major issue. Additionally, conventional methods for laying of CF impose limits on the achievable design geometries.

To the best of our knowledge, CLEVERarm is the first exoskeleton to largely incorporate CF reinforced plastic within its structure. To address the aforementioned challenges, extensive finite element analysis were done to ensure that maximum deflection of the structure in worst case loading scenarios is within the acceptable range. Plastic/metal interfaces were designed by distributing the load on larger surfaces to avoid concentration of stress [17]. Finally, using 3D printers capable of manufacturing CF reinforced plastic parts enabled achieving complex design geometries for the links of the exoskeleton.

CLEVERarm is enhanced with games to boost the engagement of patients in therapy. Integration of gaming with robotics based rehabilitation therapy has proven to be successful for therapeutic goals. CLEVERarm uses Augmented Reality (AR) technology developed by Hololens (Microsoft Corporation) to provide a different gaming experience to stroke patients. Games played on 2D displays make perception of depth in 3D difficult, and are not usually in one-to-one scale with the actual motion of the arm. Augmenting virtual holograms into the 3D space addresses both issues, and therefore AR games can be an effective and a more immersive alternative to classic game environments for patients not suffering from visual and cognitive deficits. Several game concepts based on reaching motions in 3D space have been developed and integrated into the CLEVERarm to this end. Figure 2 shows examples of such games:

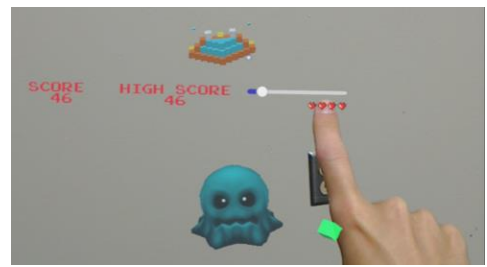

Fig 2. Games developed in Augmented Reality Environment

Game environments are part of the control architecture of the CLEVERarm since they represent the final desired position for the patient hand. As figure 3 shows, the game environment is both a display for providing visual feedback and clues to the patient, and simultaneously acts as the input for the reference generator for the exoskeleton. Reference generation block within the control architecture uses the algorithms developed by the authors for generating human-like motions considering the scapulohumeral rhythms [18, 19]. Given the desired final position for the hand, the algorithm generates a reference path for the exoskeleton in the configuration space. The reference generated by the block and the feedback on the current configuration of the robot is used for calculation of feedforward and feedback control signals to cancel the gravity and inertia of the exoskeleton and provide assistance.

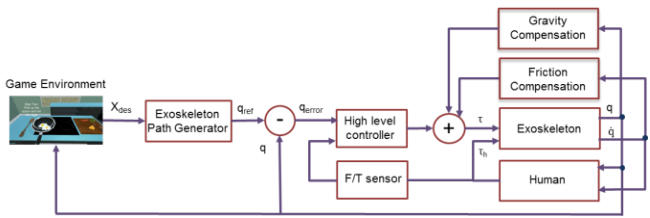

Fig 3. CLEVERarm Control Architecture

III. CHALLENGES AND OPPORTUNITIES

Development of home-based rehabilitation systems is the ultimate goal for achieving high intensity in post stoke therapy. Home-based exoskeletons for example could dramatically improve the intensity and effectiveness of therapy received by patients [20]. Robots could allow patients to start therapy in the very early stages of recovery, without having to deal with the hassles of frequent and long visits to clinics. In the comfort of their own homes, people could get specific training at the appropriate level of intensity, overseen and monitored by a human therapist over the internet [21].

Currently portability, high costs and limitations on the performance of the available systems are the main barriers for using rehab exoskeletons in patients' homes. Emergence of new technologies can help achieving lighter and more portable devices as evidence by CLEVERarm. On the other hand, cost is a more complicated issue that is influenced by the public policy and insurance industry. There is a consensual belief among the main players in the rehabilitation eco-system that the price of robotics is unrealistically high to be adapted widely. However, the results of studies show that by including the dosage of the treatment within the analysis and considering the possibility of several-to-one therapy paradigms enabled by robotic devices, the cost of using robotic systems is very close to conventional therapy [22]. Prevalence of such studies that can quantitatively demonstrate the significance of the benefits of technology based rehabilitation, can ultimately alter the stereotypes on the high cost of robotic devices.

Maximizing therapy robots' ability to help patients depends on deepening the human-robot interaction. This sort of connection is the subject of significant research of late, and not just for patient treatment. In most cases of human-robot collaboration, the human takes the lead role; however, in therapy the interaction is significantly more complicated and the robot must closely observe the patient and decide when to provide corrective input. This is signified in the efforts for designing the so-called assist as needed control paradigms for rehabilitation exoskeletons [23]. Effective integration of biologic signals such as muscle Electromyography (EMG) [24] or brain Electroencephalography (EEG) [25] within the control architecture of the exoskeleton has been central in such efforts. Despite many significant advances in this area, lack of intelligent control strategies that can realize minimal assistance paradigms is still an open problem.

Using of gaming along with robotic devices has proven to be an effective tool for rehabilitation purposes. New developments in immersive technologies such as virtual reality devices and the recently developed augmented reality systems can be adapted to be used with rehabilitation exoskeletons to enable diverse training possibilities. Although linking the real and virtual worlds within these systems is a challenging task, an exoskeleton equipped with a high fidelity virtual or augmented-reality device could offer unique experience for patients who do not suffer from vision or cognitive deficits as a result of stroke. Moreover, such technologies can help reduce the social isolation many stroke patients experience. With the aid of augmented reality tools, therapy robots can help patients interact with each other, as in a virtual exercise group. This sort of connection can make rehabilitation a pleasant experience in patients' daily lives, one they look forward to and enjoy, which will also promote their recovery. Obviously, further research and development is required for testing the effectiveness and possibility of the adaption of such methods within rehabilitation practice.

IV. CONCLUSION

This paper briefly reviews the use of exoskeletons in rehabilitation of patients with neurological impairments and the challenges involved. Key features of a recently developed exoskeleton, CLEVERarm, which addresses several shortcomings of the current devices is discussed. The paper is concluded by several remarks on the opportunities new technological developments can offer for rehabilitation of patients.